\documentclass[letterpaper, 10 pt, conference]{ieeeconf}
\IEEEoverridecommandlockouts
\usepackage{graphicx} 
\usepackage{caption}
\usepackage{subcaption}
\usepackage{adjustbox}
\usepackage{hyperref}
\usepackage[table]{xcolor}
\usepackage{efbox,graphicx}
\usepackage{array, makecell} 
\usepackage[skip=1ex, 
            font=small,labelfont=bf]{caption}
\usepackage{cite}
\usepackage{amsmath,amssymb,amsfonts}
\usepackage{eucal}
\usepackage{bm}
\usepackage{algpseudocode}
\usepackage{algorithm}
\usepackage{graphicx}
\usepackage{textcomp}
\usepackage{xcolor}
\usepackage{url}

\usepackage{amsthm}
\newtheorem{remark}{Remark}

\renewcommand{\baselinestretch}{0.93}
\graphicspath{{Figures/}}
\def\BibTeX{{\rm B\kern-.05em{\sc i\kern-.025em b}\kern-.08em
    T\kern-.1667em\lower.7ex\hbox{E}\kern-.125emX}}
\begin{document}

\title{LASMP: Language Aided Subset Sampling Based Motion Planner\\
}

\author{Saswati Bhattacharjee$^{1}$, Anirban Sinha$^{2}$, Chinwe Ekenna$^{1}$
\thanks{$^{1}$ University at Albany, Department of Computer Science, NY, USA. Email:\texttt{\{sbhattacharjee, cekenna\}@albany.edu}, $^{2}$ GE Aerospace Research, NY, USA.
Email:\texttt{\{anirban.sinha1@ge.com\}
}
}
}


\maketitle
\begin{abstract}
This paper presents the Language Aided Subset Sampling Based Motion Planner (LASMP), a system that helps mobile robots plan their movements by using natural language instructions. LASMP uses a modified version of the Rapidly Exploring Random Tree (RRT) method, which is guided by user-provided commands processed through a language model (RoBERTa). The system improves efficiency by focusing on specific areas of the robot's workspace based on these instructions, making it faster and less resource-intensive. Compared to traditional RRT methods, LASMP reduces the number of nodes needed by 55\% and cuts random sample queries by 80\%, while still generating safe, collision-free paths. Tested in both simulated and real-world environments, LASMP has shown better performance in handling complex indoor scenarios. The results highlight the potential of combining language processing with motion planning to make robot navigation more efficient.
\end{abstract}


\section{Introduction} \label{sec:introduction}
Autonomous robot navigation has expanded into numerous areas, including indoor and outdoor exploration, collaboration among multiple agents, localization and mapping (SLAM)~\cite{durrant2006simultaneous}, interaction with humans, self-driving technologies, search and rescue missions, warehouse automation, and space exploration. Advances in \textit{Visual Large Language Models} (VLLMs)\cite{anderson2018vision,deruyttere2019talk2car,codevilla2018end,tellex2011understanding, matuszek2010following} have enabled autonomous navigation  to interface with humans effectively. The fundamental idea behind VLLM-based robot navigation is to map the language and visual semantics into the robot's navigation controls in an end-to-end fashion. The mapping is done by training one or multiple neural networks in supervised\cite{shah2023lm,hu2023planning}, imitation\cite{codevilla2018end}, or reinforcement learning frameworks\cite{zhu2017target}. To train these models, several works have developed datasets~\cite{geiger2012we,anderson2018vision,deruyttere2019talk2car,sriram2019talk,rufus2021grounding}. A recent advancement in this field is to reason about the decisions made by the trained networks ~\cite{kim2020advisable,kim2019grounding} and interact with the user to make navigation decisions for the next steps with the help of \textit{Generative Pretrained Transformer} models~\cite{mao2023gpt,sha2023languagempc,ma2024lampilot}. Despite the success of VLLMs for indoor\cite{anderson2018vision,hu2019safe,shah2018follownet} and outdoor\cite{tellex2011understanding,kim2020advisable,sriram2019talk} autonomous navigation, a major limitation has been the requirement of exponentially large datasets\cite{shalev2016sample} for training.


\begin{figure}[!hbt] 
    \centering
    \includegraphics[width=0.42\textwidth]{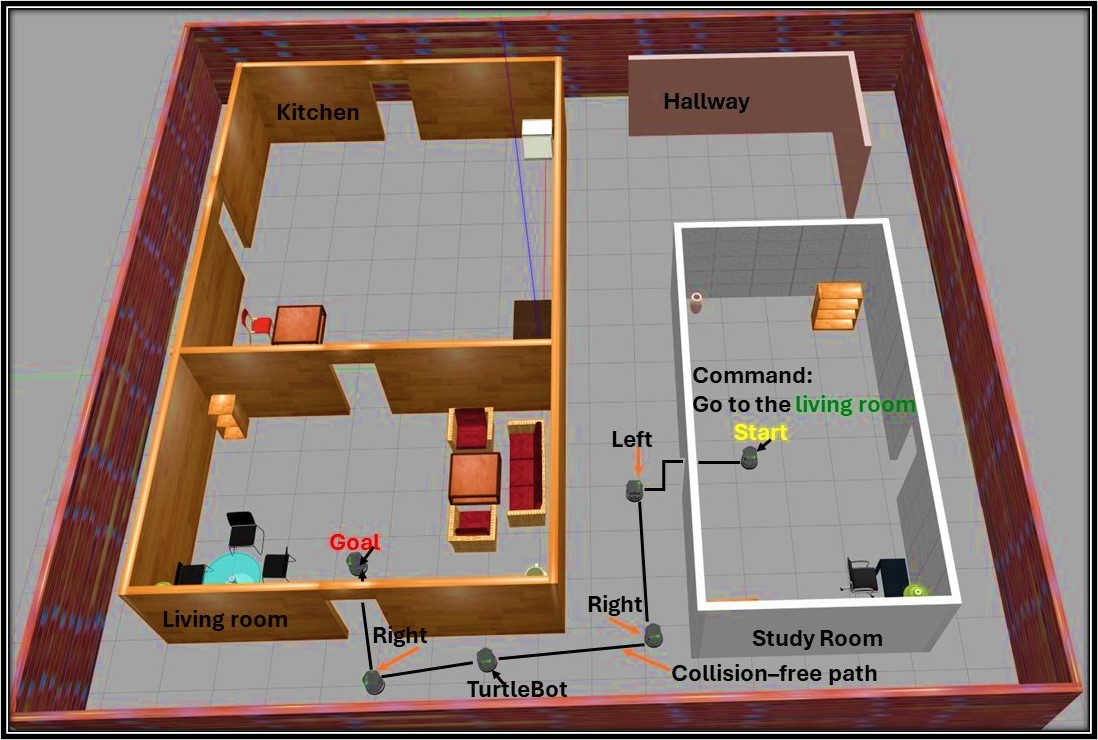}
    \caption{A robot uses LASMP to receive high-level textual or speech instructions over the cloud and parses that instruction to find a collision-free path by utilizing language grounded RRT planner. (The objects such as sofas, chairs of Figure adopted from\cite{rasouli2017effect}.)}
    \label{fig:introfigure}
\end{figure}
\begin{figure*}[!htb]
    \centering
    \includegraphics[width=0.93\textwidth]{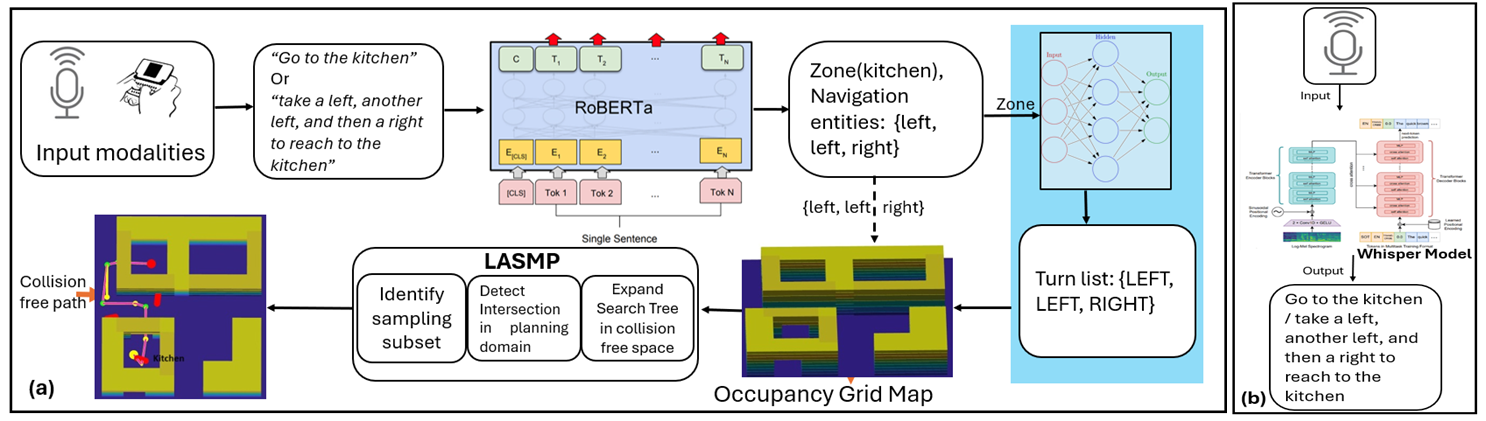}
    \caption{The RoBERTa model parses the user instruction to identify the navigation ("left", "right", etc.) and destination ("ZONE") entities. If only the "ZONE" entity is identified, it is parsed again through a neural network (blue shaded blocks) to identify the turn list. Alternatively, the turn list is directly extracted from the command and inputted into our proposed subset sampling-based planner as shown by the dashed arrow. LASMP initiates an efficient sampling-based path search by intelligently focusing on a subset of the workspace to draw valid state samples and produce a collision-free path to the goal. An extended ASR workflow,Fig.\ref{fig:flow_LASMP}(b), shows the speech to text processing pipeline.}
    \label{fig:flow_LASMP}
\end{figure*}

In many of the end-to-end robot navigation research,~\cite{hu2019safe,jain2023ground,kuo2018deep,kuo2020deep}, RRT~\cite{lavalle2001randomized, kuffner2000rrt} or its variants~\cite{karaman2011sampling,zucker2007multipartite} are used to find global or local plans. Although RRT planners are probabilistically complete, they are not sample-efficient. Specifically, random samples are often discarded because they cannot be added to the search tree, leading to unnecessary computational load on autonomous vehicles with limited energy resources. Although the work in~\cite{gammell2014informed} addresses this issue by using a prolated ellipsoidal subset of the workspace, it requires a precomputed path; otherwise, the algorithm behaves like standard RRT until a path is found.
Fig.~\ref{fig:introfigure} illustrates our proposed planner. When the robot receives the command \textit{Go to the living room}, it interprets the necessary navigation command (NC) as \textit{"left", "right", "right"} to reach the destination. We introduce the Language Aided Subset Sampling Based Motion Planner (LASMP) for mobile robot navigation, enhanced with language assistance. The overview of LASMP is shown in Fig.~\ref{fig:flow_LASMP}. Upon receiving user instructions via text or speech, a trained LLM identifies the destination and optionally navigation entities like \textit{left} or \textit{right} on a metric map\cite{kostavelis2015semantic}. If no navigation entities are provided, the robot’s current and destination positions are fed into a pre-trained network, which identifies a sequence of navigation instructions. Using these instructions, a modified, sample-efficient RRT planner, one of the paper's key contributions, computes a collision-free path.

Our contributions are
\begin{itemize}
\item LASMP provides a structured system that converts natural language instructions (both text and speech) into low-level motion commands. It is tested on different robots across various environments.
\item We introduce an improved RRT planner that leverages language-based inputs, significantly improving sample efficiency. The planner dynamically adjusts the sampling area based on the robot’s current pose and the user's commands, leading to faster and more efficient path planning compared to traditional RRT approaches.
\item We developed a dataset that includes destination names or navigation-related instructions. This dataset was used to train a transformer-based model (RoBERTa) to improve the system’s ability to recognize navigation entities and plan paths accordingly.
\item LASMP combines natural language processing and path planning by interpreting abstract high-level commands into precise turn-by-turn directions. This not only enhances the robustness of the system but also ensures that it can handle varying user input styles.

\end{itemize}



\section{RELATED WORK} \label{sec:related_work}

Language commands have shown great potential in guiding autonomous vehicles (AVs) for human-robot interaction. Several approaches have been developed to ground natural language commands into controls for AVs. A popular method is \textit{end-to-end} navigation, where language, vision, or other sensor data are fused to train AI models for generating motion commands. In~\cite{deruyttere2019talk2car,rufus2021grounding}, language instructions are combined with local vision data to train networks for sequential control inputs, while~\cite{mei2016listen} trains a LSTM-RNN to translate language commands into motion sequences. A modular approach is proposed in~\cite{sriram2019talk}, where a natural language encoder and semantic image input suggest potential waypoints, and an RRT planner generates a collision-free path. Room-to-Room (R2R) navigation using vision and language has been explored in~\cite{anderson2018vision,shah2018follownet}. While \textit{end-to-end} methods have shown success, they often require large labeled datasets, limiting their use.

Recent advancements in large language models (LLMs) integrate multiple modalities, such as audio~\cite{gong2023listen, zhang2023speechgpt}, video~\cite{xu2024drivegpt4, chen2023videollm}, and point cloud data~\cite{guo2023point, xu2023pointllm}. Building on these works, we employed a speech recognition approach to interpret spoken language and enhance model diversity.

Another line of research uses natural language to define constraints in robot navigation. In~\cite{hu2019safe}, a local cost map is updated based on language instructions mapped to environmental objects. Generalized Grounded Graphs ($\bold{G}^3$)\cite{kollar2013generalized} and Dynamic Grounding Graphs (DGG)\cite{park2017generating} dynamically parse language to identify motion constraints. Similarly,\cite{howard2014natural} presents motion planning as a constrained optimization problem, where constraints activate or deactivate based on language input. However, extracting complex constraints from language remains challenging.

Recent work like GPT-Driver~\cite{mao2023gpt} treats AV motion planning as a language modeling problem, mapping trajectory data into words for safe navigation. In~\cite{chen2024driving}, LLMs fuse environmental data to generate AV controls with reasoning, while DriveGPT4~\cite{xu2024drivegpt4} interprets video sequences for control signals. Further, LLMs can convert scene and guidance data into numerical instructions for controllers, as seen in~\cite{sha2023languagempc}, and LaMPilot~\cite{ma2024lampilot} autonomously generates code to enhance AV functionality. However, most of these approaches are still in the simulation stage with limited real-world testing.

Language models have also been applied to aid plan for manipulation tasks. In~\cite{xie2023language}, a framework learns a collision function based on robot state, scene information, and language prompts, using it for motion planning. \textit{DeepRRT}~\cite{kuo2018deep, kuo2020deep} combines vision data, language lexicons, and a proposal layer to guide RRT planners in exploring robot configurations. However, none of these methods integrate language input to improve the sample efficiency of RRT planners.

Our method uses NCs to make RRT planners sample efficient for generating low-level controls. We introduce a novel modified RRT algorithm that leverages language instructions to dynamically select a subset of the robot’s workspace for random sampling to grow the search tree. Unlike Informed RRT$^*$~\cite{gammell2014informed}, which requires a precomputed path and adjusts only the volume of the subset, our method intelligently adjusts both the orientation and size of the subset based on language input, without needing a precomputed path. Additionally, while~\cite{kuo2020deep} focuses on optimizing node extension direction using vision data, which requires significant computational resources, our approach improves sample efficiency using a ray-casting technique to sense the environment, making it an energy-efficient solution.


\section{METHODOLOGY}\label{sec:methodology}
Fig.~\ref{fig:flow_LASMP}(a) shows the workflow of computing a collision-free navigation path from textual or verbal commands, while Fig.~\ref{fig:flow_LASMP}(b) extends it by illustrating the transcription of verbal commands using the Whisper architecture \cite{radford2023robust}. The proposed robot navigation framework consists of two main modules. As described in section~\ref{sec:language_model}, the first module converts high-level natural language instructions into low-level NCs. For example, the instruction \textit{"go to kitchen"} is transformed into the sequence $\left[ left \rightarrow right \rightarrow left\right]$ with identified destination \textit{kitchen}. The second module, LASMP, is a sample-efficient planner that uses these low-level commands to guide the path search, as detailed in section~\ref{sec:LASMP_main_method}.

\subsection{Grounding Natural Language Instructions} \label{sec:language_model}
The proposed framework handles speech-based instructions with a transformer-based Automatic Speech Recognition (ASR) model 'Whisper' \cite{radford2023robust}, to convert spoken commands into texts. Whisper processes the audio by dividing it into 30-second segments, converting these into log-Mel spectrograms, and then encoding and decoding them to produce text. This text is then analyzed using RoBERTa in the NER pipeline to extract location information along with the NCs. The motion commands, locations, and their mappings with our seven unique entities are detailed in Table \ref{tab:Tablecommnads}.  For the NER task, we utilized the \textit{en\_core\_web\_lg} model \cite{Honnibal_spaCy_Industrial-strength_Natural_2020} and two transformer models, BERT \cite{devlin2018bert} and RoBERTa \cite{liu2019roberta}. As shown in Fig.~\ref{fig:figbar}, RoBERTa, based on BERT's architecture, performed best in entity extraction due to its bidirectional transformer design, which effectively captures contextual information. 

The output of the NER task is either the combination of turns and destination or it could be destination only. For the prior, the coordinates of the identified destination are retrieved from a predefined look-up table. If only the destination is outputted from the NER then the associated turns are obtained through a pre-trained feedforward network (blue-shaded module in Fig.\ref{fig:flow_LASMP}(a)) which maps the concatenated vector consisting of the coordinates of the robot's current position and coordinates of the destination to a class associated with a specific turn list. For our implementation, the network is trained for classifying a maximum of four turns which is sufficient for many indoor environments. However with appropriate data, the network can be easily trained for more turns.
\setlength{\belowcaptionskip}{-5pt}
\begin{table}[!thb]
\caption{Diverse instructions mapped into unified entities.}
\begin{center}
\begin{tabular}{ |c|c| } 
 \hline
 \rowcolor{lightgray}  \textbf{Diverse Commands} & \textbf{Unified Instruction} \\ \hline
 \makecell{go straight, move straight \\go ahead, proceed in a straight line } & Straight \\ \hline
 \makecell{go right, turn right, move right, \\take a right, right turn, go rightward} & Right \\ \hline
 \makecell{turn left, left turn, take a left,\\ move left, head left, go leftward} & Left \\ \hline
\makecell{go down, move down, go back,} & Backward \\ \hline
 \makecell{do not/avoid taking/skip/not \\to take (turns (Right/Left))} & NR / NL\\ \hline
\makecell{bedroom, kitchen, living room \\ dining room, bathroom, laundry} & ZONE\\ \hline
\end{tabular}
\label{tab:Tablecommnads}
\end{center}
\end{table}
\vspace{-0.5cm}

\subsection{Language Assisted Sampling-Based Motion Planner}\label{sec:LASMP_main_method}
Let's define the planning domain as $\mathcal{D}\subset\mathbb{R}^n$ and $\mathcal{D}_{obs}\subset\mathcal{D}$ is the set of states in collision. Then the set of collision-free states is $\mathcal{D}_{free}=\mathcal{D}\setminus\mathcal{D}_{obs}$. An element $\bm{x}(t)\in \mathcal{D}$ denotes the state of the robot at time instant $t$. Let $\Phi=\left[\phi_1, \dots, \phi_n \right]$ denote an ordered list of the low-level NC inferred from the input textual \textit{(or speech)} instruction with its elements as $\phi_i, i\in\left\{1,\dots, n\right\}$. Let's also define the augmented state of the robot as $\bm{\bar{x}}(t) = \left(\bm{x}(t),\phi_i\right)$. The operator $(\cdot)^\vee$ when applied to the augmented state returns the actual state, \textit{i.e.,} $\bm{\bar{x}}^\vee(t)\rightarrow \bm{x}(t)$. Given the start and goal state of the robot as $\bm{\bar{x}}_s$ and $\bm{\bar{x}}_g$ and NC list $\Phi$, the LASMP finds a collision-free path
\begin{equation}
    \mathcal{P} = \left\{\bm{\Bar{x}}(t)|\bm{\bar{x}}(0)=\bm{\bar{x}}_s, \bm{\bar{x}}(t_f)=\bm{\bar{x}}_f, \forall t \, \bm{\bar{x}}^\vee(t)\in\mathcal{D}_{free}\right\}
\end{equation}
For brevity, we will drop $t$ to express the robot's state for the remaining of the paper or specify it as needed.

\textbf{LASMP Solution Methodology: }
The LASMP is a sample-efficient RRT-type motion planner that can find a collision-free path by sampling from a smaller subset of the planning domain $\mathcal{D}$. Since the planner is informed with an ordered turn list $\Phi$, the planner employs a local search strategy to find a collision-free path to the intersections to complete the turns in order one by one. This structure of the problem allows us to decompose the planning problem into several smaller subproblems. Given a planning problem $\mathcal{P}$ and an associated $\Phi$ with $n$ commands, we need to solve $(n+1)$ planning sub-problems. Mathematically we can write,
\begin{equation}
    \mathcal{P} = \left\{\mathcal{P}_0, \dots \mathcal{P}_{n+1}\right\} \equiv \left\{\mathcal{P}_j\right\} \quad \text{where} \quad j \in \left\{0, \dots, n+1\right\}
\end{equation}
The goal state of the path for $\mathcal{P}_{j}$ becomes the initial state of the sub-problem $\mathcal{P}_{j+1}$ with the NC updated to $\phi_{j+1}$ \textit{i.e.},
\begin{eqnarray}
    \label{eq:plan_subprobs_series}
    \mathcal{P}_{j} &=& \left\{\bm{\bar{x}}(t_k), \dots, (\bm{\bar{x}}^\vee(t_{k+m}), \phi_{j})\right\} \\
    \mathcal{P}_{j+1} &=& \left\{(\bm{\bar{x}}^\vee(t_{k+m}),\phi_{j+1}), \dots, \bm{\bar{x}}(t_{k+m+r})\right\}
\end{eqnarray}
where $m$ and $r$ are the number of states in the paths for $\mathcal{P}_j$ and $\mathcal{P}_{j+1}$. Next solution method for $\mathcal{P}_{j}$ is discussed.

\textbf{Solution methodology of $\mathcal{P}_{j}$:} The planning sub-problems, $\mathcal{P}_j$s, only have the starting states defined whereas their goal states are not known uniquely before the problems are solved, except for $\mathcal{P}_{n+1}$ which has its goal state as $\bm{x}_f$. However, since $\phi_j$ is known, the goal state $\bm{x}_{jf}$ of the planning problem $\mathcal{P}_j$ could be any element of the set $\mathcal{D}_{is}^j \subset \mathcal{D}$, \textit{i.e.}, $\bm{x}_{jf} \in \mathcal{D}_{is}^j$ where $\mathcal{D}_{is}^j$ is the set associated with the states at an intersection region in the planning environment where the motion command $\phi_j$ needs to be executed. However, the region of the $\mathcal{D}_{is}^j$ is not known during the planning time either. Therefore the planning problem $\mathcal{P}_{j}$ reduces to finding a collision-free path while simultaneously identifying a goal state $\bm{x}_{jf}$ as described next.

\subsubsection{Collision-free Path for planning sub-problem $\mathcal{P}_j$}
\label{sec:collision_free_path_subset}
From now on, we will consider the planning domain $\mathcal{D} \subset \mathbb{R}^2$ for the ease of explanation of the method. Then the states and augmented states are now defined as $\bm{x}=(x_r, y_r)$ and $\bm{\bar{x}}=(x_r, y_r, \phi_j)$ respectively where $x_r,y_r$ defines the robot's position. Also for mobile robots, $\phi_j$ can be mapped to a unit direction vector, $\bm{v}_j=\left[v_x,v_y\right]^T\in\mathbb{R}^2$ towards which the robot needs to take the turn from the state $
\bm{x}_{jf}$.

Here we assume $\bm{x}_{jf}$ is known and in the next subsection, we provide method to find it. Known $\bm{x}_{jf}$ turns $\mathcal{P}_j$ into a regular planning problem but grounded with $\phi_j$. Thus instead of sampling from the whole planning domain, it will be efficient to sample from a subset of the planning domain. For user-defined parameters $h$ and $w$, we propose a rectangular subset to draw random samples to grow the search tree. The rectangular subset is a function of robot state $\bm{x}$ corresponding to the node that is being expanded to grow the search tree. The vertices of the rectangular subset are defined as
\begin{eqnarray} \nonumber
\label{eq:subset_vertex}
    \bm{x}_{v1}, \bm{x}_{v3} &=& \bm{x} \pm [w/2, h/2]^T \\
    \bm{x}_{v2}, \bm{x}_{v4} &=& \bm{x} \pm [w/2, -h/2]^T
\end{eqnarray}
Then the sampling subset is defined as,
\begin{equation} \label{eq:subset_definition}
    \mathcal{D}_{smp} = \{(x, y) | \bm{x}_{min} \leq \bm{x}_{vi} \leq \bm{x}_{max}\}, \, i \in \{1,\dots,4\}
\end{equation}
where $\bm{x}_{{min}} = (x_{min}, y_{min})$, $\bm{x}_{max} = (x_{max}, y_{max})$. Note that, in sampling-based path planning configuration space is generally preferred~\cite{lavalle2006planning,lozano1990spatial, anshelevich2000deformable} but for LASMP task space sampling~\cite{sinha2023oc3} is adopted since subset is defined in there.

\begin{remark}
Generating random samples from a local subset as compared to the complete planning domain increases the probability of sampling from the desired region by a factor of $\frac{\bm{\zeta}(\mathcal{D})}{\bm{\zeta}(\mathcal{D}_{smp})}$ where $\bm{\zeta}$ is a set measure.
\end{remark}

\begin{remark}
Sampling from a subset instead of the whole planning domain helps the planner to focus on the informed regions where openings to take turns would exist. This allows more samples to be generated in the area of interest and potentially get added to the tree. LASMP uses subset sampling to find collision-free paths and detect narrow openings where commanded turns can be executed.
\end{remark}

\subsubsection{Detecting intersection}\label{sec:planning_model_find_opening}
In section~\ref{sec:collision_free_path_subset} we assumed $\bm{x}_{jf}$ is known, but $\bm{x}_{jf}$ could be any state in the set $\mathcal{D}_{is}$. Note that $\bm{x}_{jf}$ is the terminal state for the planning sub-problem $\mathcal{P}_j$ at which the navigation instruction $\phi_j$ is executed. This implies that at the state $\bm{x}_{jf}$ if a ray is cast along the direction $\bm{v}_j$ then it will not hit any obstacle at least for a threshold distance $d$. $d$ is a user defined parameter.

In order to detect whether any state in $\mathcal{D}_{is}$ is reached while expanding a node $N$ with state $\bm{x}_{near}$ towards a randomly sampled state $\bm{x}_{rand}$, a ray is cast along the $\bm{v}_j$ from each of the discretized states between $\bm{x}_{near}$ and $\bm{x}_{rand}$. If the discretization is done with a step size $\delta$, then the $k^{th}$ intermediate state $\bm{x}^{im}_{k}$ between $\bm{x}_{near}$ and $\bm{x}_{rand}$ will be
\begin{equation}
    \bm{x}^{im}_{k} = \bm{x}_{near} + k\delta(\bm{x}_{rand} - \bm{x}_{near})/\|\bm{x}_{rand} - \bm{x}_{near}\|
\end{equation}
If for consecutive $n_{cons}$ points, \textit{i.e.,} $\bm{x}^{im}_{k}, \dots, \bm{x}^{im}_{k+n_{cons}}$, there are no obstacles found upto a distance $d$ along the direction $\bm{v}_j$, then the state $\bm{x}^{im}_{k+n_{cons}}$ is marked as the terminal state \textit{i.e., } $\bm{x}_{jf}$ of the planning sub-problem $\mathcal{P}_j$.

In Algorithm\ref{alg:lasmp_algo}, the steps of the LASMP are shown. The \textit{GetSubset} function is an overloaded function that uses Eq.\eqref{eq:subset_vertex}, while the function \textit{Intersection} implements the method in section\ref{sec:planning_model_find_opening}.  The variable $F_{turn}$ is a flag which raised to \textit{true} if $\bm{x}^{im}_{k+n_{cons}}$ is $\bm{x}_{jf}$.

\begin{algorithm}[!htb]
\caption{LASMP}
\begin{algorithmic}[1]
\State \textbf{Input}: $\bm{x}_{s}$, $\bm{x}_{f}$, $LangCue$
\State $Tree$  $\gets$ Init$(\bm{x}_{s})$
\State $\bm{x} \gets \bm{x}_{s}$
\State $\Phi \gets$ \textcolor{blue}{NavSeqFromText}($LangCue$)
\State $\bm{v}_r \gets \Phi.pop()$
\State $\mathcal{D}_{smp} \gets$ \textcolor{blue}{GetSubset}($\bm{x}$, $h$, $w$, $\bm{v}_r$) 
\While{Goal not reached}
    \State $\bm{x}_{rand} \gets$ SampleRandomSt($\mathcal{D}_{smp}$)
    \State $\bm{x}_{near} \gets$ FindNearestNode($Tree$, $\bm{x}_{rand}$)
    \If{motionValid($\bm{x}_{rand}, \bm{x}_{near}$)}
        \State $\bm{x}_{new}, F_{turn} \gets$ \textcolor{blue}{Intersection}($\bm{x}_{near}$, $\bm{x}_{rand}$, $\bm{v}_r$)
        \If{$F_{turn}$}
            \State $\bm{v}_r \gets \Phi.pop()$
            \State $\mathcal{D}_{smp} \gets$ \textcolor{blue}{GetSubset}($\bm{x}_{new}$, $h$, $w$, $\bm{v}_r$)
        \EndIf
        \State $Tree \gets$ appendNode($\bm{x}_{new}$)
        \State $Tree \gets$ appendVertex($\bm{x}_{new}$, $\bm{x}_{near}$)
        \If{IsGoal($\bm{x}_{new}$)}
            \State break
        \EndIf
    \EndIf
\EndWhile
\State $path \gets$ ExtractPath($Tree$, $\bm{x}_{s}$, $\bm{x}_{g}$)
\State \Return $path$
\end{algorithmic}
\label{alg:lasmp_algo}
\end{algorithm}

\section{Implementation Details}\label{sec:experiments}
LASMP  is tested in Coppeliasim~\cite{6696520} environment using the MATLAB remote interface on an Ubuntu machine with an Intel i7 processor, 16GB RAM, and an NVIDIA GeForce RTX GPU. We used four planning scenes—three simulated (Fig.~\ref{fig:House}) and one real-world (Fig.~\ref{fig:real_world})—with three robots, Turtlebot3 ~\cite{turtlebot} and Pioneer3DX ~\cite{pioneer_robot} for simulation, and Turtlebot2~\cite{turtlebot} for real-world experiments. The SpaCyV3~\cite{Honnibal_spaCy_Industrial-strength_Natural_2020} framework was used to train the language models.

\subsection{Planning Scenarios}
\begin{figure}[!htb]
    \centering
    \includegraphics[width=0.5\textwidth]{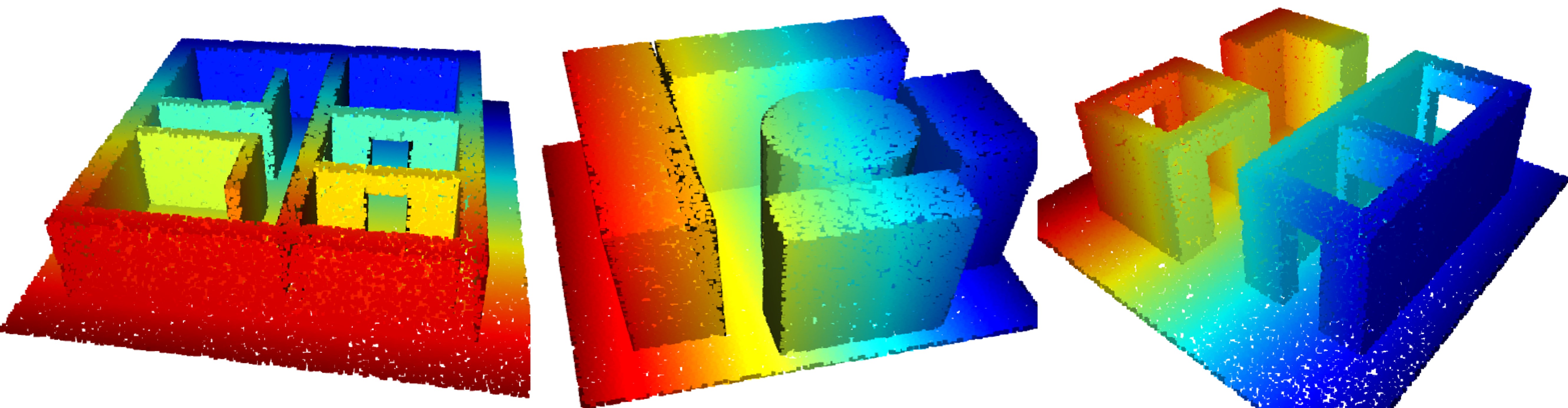}
    \caption{3D occupancy grids of the planning scenarios for evaluating the effectiveness of the LASMP: (left) office space (OS) (middle) random obstacle scene (RO) and (right) domestic environment (DE).}
    \label{fig:House}
\end{figure}

The three planning scenes considered in the paper are the domestic environment (DE), office space (OS), and random obstacle (RO) scenes. The point cloud representations of these scenes are converted into three-dimensional occupancy grid maps for planning purposes as shown in Fig.~\ref{fig:House}. The paths computed by LASMP are smoothened before executing on the robots using a simple pure pursuit-type controller. For safe obstacle avoidance, inflated occupancy grid maps are used. The test parameters for LASMP are detailed in \cite{the_dataset}.

For all the planning scenarios, the performance of LASMP is evaluated against classical RRT with the following metrics (i) the number of nodes added to the search tree and (ii) the number of queries to the random sample generator.

\subsection{Training Language Models:}
\begin{figure}[!htb]
\centerline{\includegraphics[width=0.42\textwidth]{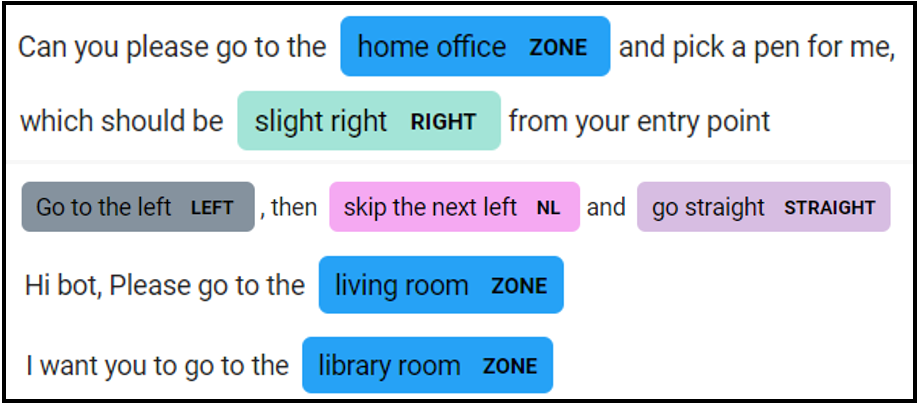}}
\caption{Performance of RoBERTa model trained on our dataset in predicting navigation and destination entities from instructions.}
\label{fig: displacy}
\end{figure}

\textbf{Dataset}: The training dataset\footnote{ dataset is made available at 
 \textcolor{blue}{https://github.com/LASMP23/LASMP}} includes 350 motion-related commands and location phrases. Generated using the GPT-4 model\cite{openai2024chatgpt}, the dataset features both simple prompts (e.g., “Move to [A / B etc] location”) and complex prompts (e.g., “Take a [left/right] turn, then turn [left/right] to reach the goal”). It was annotated with an annotation tool \cite{spacy_annotation_tool} and saved in JSON format for NER task. For example, \textit{“Move forward to the music room”} is annotated as \textit{(20,30,“ZONE”)}, where the first two elements denote the start and end indices of the entity and the third element is the type. We also varied command phrasing in the dataset (e.g., “take a right,” “move right”) for enhanced performance of RoBERTa (see Table \ref{tab:Tablecommnads}). The dataset was divided into training and test sets with a 4:1 ratio. Data preparation involved creating spaCy DocBin objects from annotated data, initializing the configuration file using the spaCy Command Line Interface, and training the language model. The results are shown in Figures~\ref{fig:figbar}(b) and~\ref{fig: displacy}.

\textbf{Transformer Architectures}: We used Whisper \cite{radford2023robust} model for ASR, based on the standard encoder-decoder transformer architecture \cite{vaswani2017attention}, to transcribe speeches into texts. RoBERTa \cite{liu2019roberta} was fine-tuned for the NER task.

\section{Results}\label{sec:results}
In section~\ref{sec:performance_language_models} we first describe the performance of the transformer-based models in predicting navigation-related entities from the speech and textual commands. In section~\ref{sec:performance_LASMP} we evaluate the performance of LASMP in finding paths as compared to classical RRT, followed by experimental results.

\subsection{Evaluation of speech and language models} \label{sec:performance_language_models}
\begin{figure}
    \centering
    \includegraphics[width=0.4\textwidth]{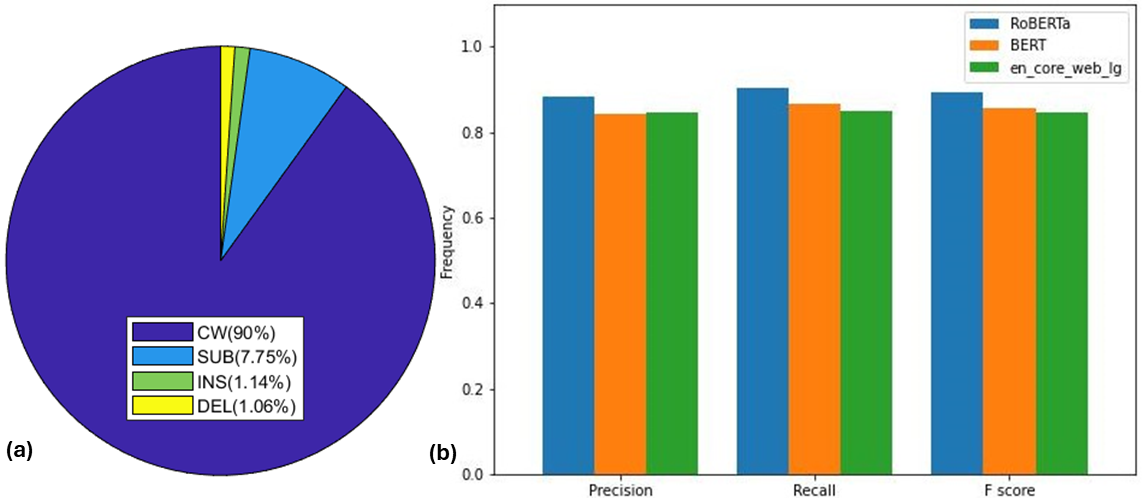}
    \caption{Performance of (a)Whisper model for transcribing speech to text (b)the different language models for NER task. }
    \label{fig:figbar}
\end{figure}

This section presents the performance analysis of the ASR and NER tasks. The Whisper's performance for ASR task was evaluated using several metrics: Number of Correct Words (CW), Number of Deleted Words (DEL), Number of Substituted Words (SUB), Number of Inserted Words (INS), Word Error Rate (WER), and Character Error Rate (CER). Fig. \ref{fig:figbar}(a) illustrates the performance of the Whisper model which is obtained by utilizing a subset of our dataset. The results in Fig. \ref{fig:figbar}(a) show that the Whisper model correctly identified $90\%$ of the words. We achieved an average error rate for WER of $0.08728$ and for CER of $0.0487$, indicating good transcription accuracy in both word and character-level metrics. To evaluate the performance of NER task, we trained language models using our dataset, and the resulting performance metrics were presented in Table\ref{tab:Table2} and an illustration was presented in Fig.\ref{fig:figbar}(b). The results demonstrate the strong performance of RoBERTa in identifying the custom entities, with high precision (0.883), recall (0.903), and
\begin{table}[!thb]
\begin{center}
\caption{Performance evaluation of the language models.}
\begin{tabular}{ |c|c|c|c| } 
 \hline
 \rowcolor{lightgray} \textbf{Language Models} & \textbf{Precision} & \textbf{Recall} & \textbf{F score} \\ \hline
 RoBerta & \makecell{\textbf{0.883}}&\makecell{\textbf{0.903}} &\makecell{\textbf{0.893}} \\ \hline
 BERT & \makecell{0.842} &\makecell{0.868} &\makecell{0.855} \\ \hline
 en\_core\_web\_lg & \makecell{ 0.845} &\makecell{ 0.851} &\makecell{ 0.848} \\ \hline
 \end{tabular}
\label{tab:Table2}
\end{center}
\end{table}
 F1 score (0.893) which ensures transformer-based models are more suitable for our NER task than deep learning-based models. 
\vspace{-0.25 cm}
\vspace{-0.05 cm}
\begin{figure}[!htb]
    \centering
    \includegraphics [width=0.47\textwidth]{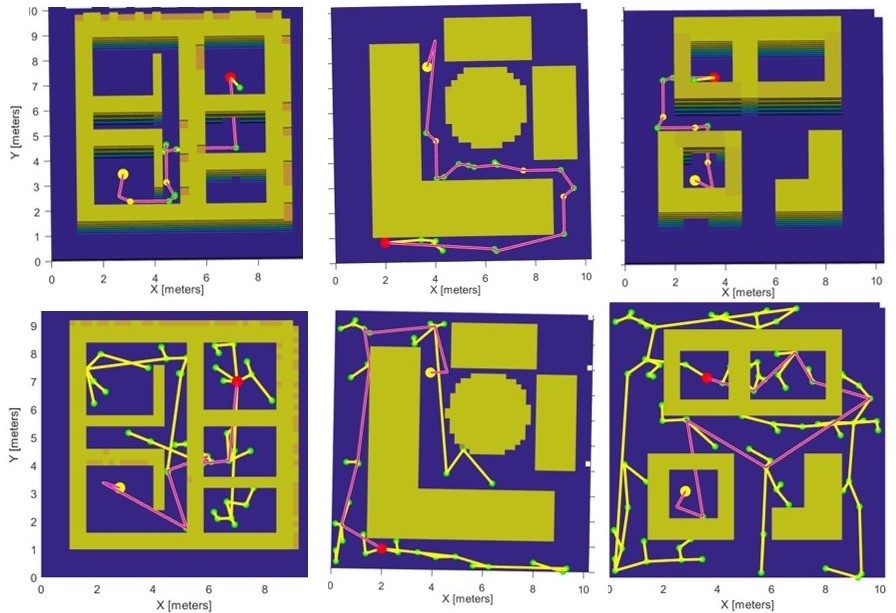}
    
    \caption{Comparison of the feasible paths computed using LASMP (top row) and RRT(bottom row) in the three distinct planning scenes as shown in Figure~\ref{fig:House}. The paths are highlighted in purple. The green circles and yellow straight lines depicted the vertices and edges of the search trees.The red and yellow circles represent start and goal.}
    \label{fig:simulationenv}
\end{figure}
\begin{figure}[!htb]
    \centering
   \includegraphics[width=0.5\textwidth]{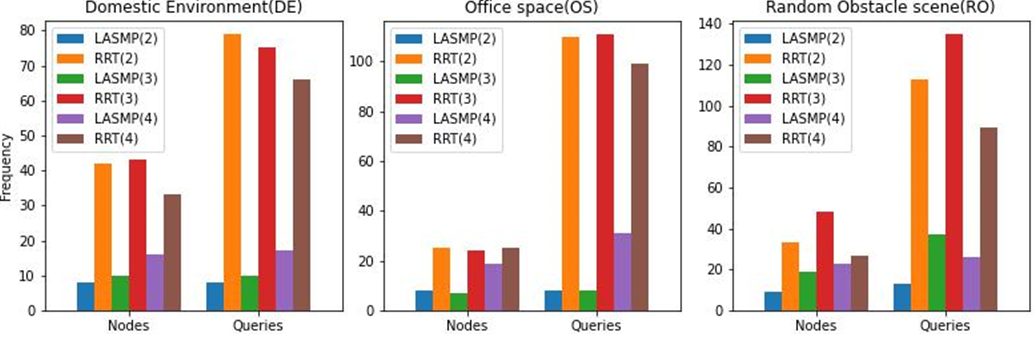}
   \caption{Performance comparison of LASMP with RRT concerning the number of nodes to find the path and queries made to the random state generator function. LASMP(\#) represents the number of turning instructions provided in the language cues.}
   \label{fig:RRT_compare}
\end{figure}

\begin{table}[!htb]
\begin{center}
\caption{Comparison of LASMP and RRT with respect to the Path Lengths (PL) and Elapsed Time (ET) on various scenarios.
}
\resizebox{\columnwidth}{!}{%
\begin{tabular}{ |c|c|c|c|c| } 
 \hline
\rowcolor{lightgray} \textbf{Environment} & \textbf{PL LASMP} [m] & \textbf{PL RRT} [m] &\textbf{ET LASMP(sec.)} &\textbf{ET RRT(sec.)} \\ \hline
 DE & \textbf{9.4535} & 17.1151 & \textbf{4.9796}  & 76.9201\\ \hline
 OS & \textbf{11.2680} & 12.7969 & \textbf{7.2372} & 96.72\\ \hline
 RO & 22.2368 & \textbf{14.4388} & \textbf{21.2200} & 194.3100\\ \hline 
 UBC & \textbf{16.0491} & 24.7520 & \textbf{16.0657} & 452.5521\\ \hline
\end{tabular}
}
\label{tab:path_length}
\end{center}
\end{table}

\begin{table}[!htb]
\caption{Comparison of the number of nodes added to the search tree and the number of queries to the random state generator function during the path search for the LASMP and RRT planners.}
\Huge
\begin{adjustbox}{width=\columnwidth,center}
\begin{tabular}{|c|c|c|c|c|c|c|c|c|}
\hline
 \rowcolor{lightgray}& \multicolumn{2}{c| }{\textbf{LASMP\#2}} & \multicolumn{2}{c| }{\textbf{RRT\#2}} & \multicolumn{2}{c| }{\textbf{LASMP\#3}} & \multicolumn{2}{c| }{\textbf{RRT\#3}} \\
\hline
\rowcolor{lightgray}& \#Nodes & \#Queries & \#Nodes & \#Queries & \#Nodes & \#Queries & \#Nodes & \#Queries\\
\hline
DE & \textbf{8} & \textbf{8} & 42 & 79 & \textbf{10} & \textbf{10} & 43 & 75 \\
\hline
OS & \textbf{8} & \textbf{8} & 25 & 110 & \textbf{7} & \textbf{8} & 24 & 111\\
\hline
RO & \textbf{9} & \textbf{13} & 33 & 113 & \textbf{19} & \textbf{37} & 48 & 135 \\
\hline
\end{tabular}
\end{adjustbox}
\label{tab:comparison}
\end{table}
\begin{table}[!htb]
\caption{This table is an extension of Table\ref{tab:comparison_turn3}}
\Huge
\begin{adjustbox}{width=0.6\columnwidth,center}
\begin{tabular}{|c|c|c|c|c|}
\hline
 \rowcolor{lightgray}& \multicolumn{2}{c| }{\textbf{LASMP\#4}} & \multicolumn{2}{c| }{\textbf{RRT\#4}} \\
\hline
\rowcolor{lightgray}& \#Nodes & \#Queries & \#Nodes & \#Queries \\
\hline
DE & \textbf{16} & \textbf{17} & 33 & 66  \\
\hline
OS & \textbf{19} & \textbf{31} & 25 & 99 \\
\hline
RO & \textbf{23} & \textbf{26} & 27 & 89  \\
\hline
\end{tabular}
\end{adjustbox}
\label{tab:comparison_turn3}
\end{table}
\vspace{-0.05 cm}
\begin{table}[!htb]
\begin{center}
\caption{Start and goal states (position[m] and yaw[rad]) of the robot for the planned paths in Figure~\ref{fig:simulationenv}. The final column is the sequence of the navigation commands retrieved using the RoBERTa.}
\resizebox{\columnwidth}{!}{%
\begin{tabular}{ |c|c|c|c| } 
 \hline
 \rowcolor{lightgray} \textbf{Environments} & \textbf{Start} & \textbf{Goal} & \textbf{Turnlist} \\ \hline
 DE & \makecell{[3.6, 7.4, $\pi$]} &\makecell{[2.8, 3.2, $\frac{\pi}{2}$]} &\makecell{{left, left, right}} \\ \hline
 OS & \makecell{[7.0, 7.0, $-\frac{\pi}{2}$]} &\makecell{[2.8, 3.2, $\frac{\pi}{2}$]} &\makecell{{right, left, right, left}} \\ \hline
 RO & \makecell{[2.0, 0.8, 0]} &\makecell{[3.8, 7.8, $\frac{\pi}{2}$]} &\makecell{{left, left, nr, right}}  \\ \hline
 UBC & \makecell{[8.5, 2.0, $\frac{\pi}{2}$]} &\makecell{[0.8, 5.8, $-{\pi}$]} &\makecell{left, left, right} \\ \hline
\end{tabular}
}
\label{tab:simulation_example_parameters}
\end{center}
\end{table}
\subsection{Performance of the LASMP} \label{sec:performance_LASMP}

\subsubsection{\textbf{Evaluation Metrics and Performance Comparison}}
We evaluated LASMP's performance using two key metrics: (a) the number of queries made to the random function generator, and (b) the total number of nodes added to the final search tree before finding a collision-free path. Unlike optimal RRT variants, which require a precomputed feasible path, LASMP operates without such requirements, making it more flexible in real-time applications.
For the evaluation, we considered several start and goal states in each of the three planning scenes (Fig.~\ref{fig:House}). These states were selected to involve 2, 3, or 4 turns, increasing the complexity of the planning problem. Table~\ref{tab:comparison} summarizes the results across different scenes, with visual representations shown in Fig.~\ref{fig:simulationenv}. 

In the DE scene with two turning commands, RRT required 42 nodes, about 5 times more than LASMP, which only required 8 nodes. RRT also queried the random state generator 79 times, while LASMP made only 8 queries. In the more complex scenario with three turns, RRT required 43 nodes and 75 queries, compared to LASMP’s 10 nodes and 10 queries. Similarly, in the four-turn DE environment, RRT needed 33 nodes and 66 queries, while LASMP required only 16 nodes and 17 queries. These trends are consistent across other environments, as LASMP consistently outperformed RRT in both node generation and query count. On average, LASMP reduced node generation by 55\% and query generation by 80\%, demonstrating its high sample efficiency. These results, averaged over 10 independent runs, are summarized in Table ~\ref{tab:comparison} and Table~\ref{tab:comparison_turn3} and visualized in Fig.~\ref{fig:RRT_compare}, show LASMP’s superior performance across all cases.

\subsubsection{\textbf{Path Length and Real-World Execution}}
Table~\ref{tab:path_length} reports the path lengths (PL) and elapsed times (ET) for the planning problems on different environments.
In most cases, LASMP computed shorter paths than RRT. In the RO environment, LASMP took $\sim89\%$ less time to compute path compared to RRT, underscoring LASMP's time efficiency.

\begin{figure}[!htb]
    \centering
    \includegraphics[width=0.48\textwidth]{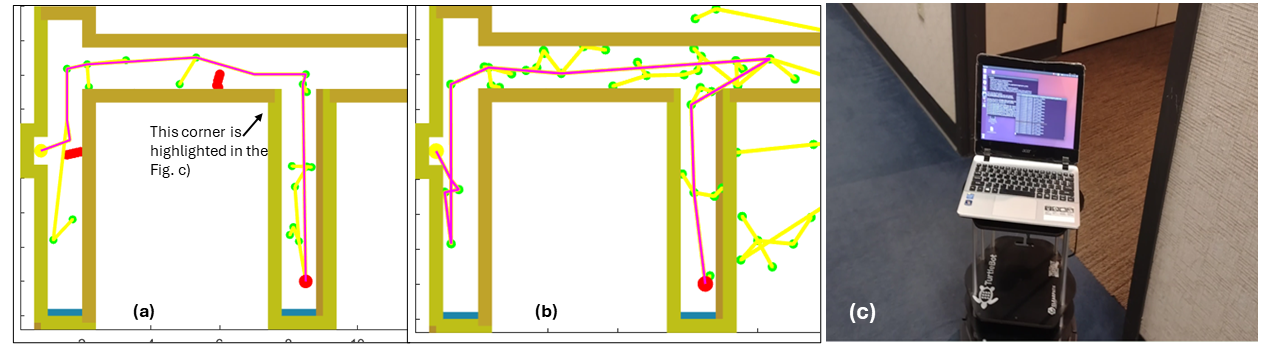}
    \caption{University building corridor environment: (a) LASMP and (b) RRT generated paths in red lines. (c) Real-world experiment.}
    \label{fig:real_world}
\end{figure}

The search trees generated by LASMP and RRT for three planning problems in three different scenes are shown in Fig.~\ref{fig:simulationenv}. The green vertices represent the nodes, and yellow edges represent the connections, with the computed paths highlighted in purple. The top row shows the search trees for LASMP, while the bottom row shows those for RRT. As evident from these visuals, LASMP required fewer nodes to find a feasible path, resulting in more efficient planning. The start and goal states, along with the retrieved navigation commands, are listed in Table~\ref{tab:simulation_example_parameters}.
\begin{figure}[!htb]
    \centering
        \includegraphics[width=0.48\textwidth]{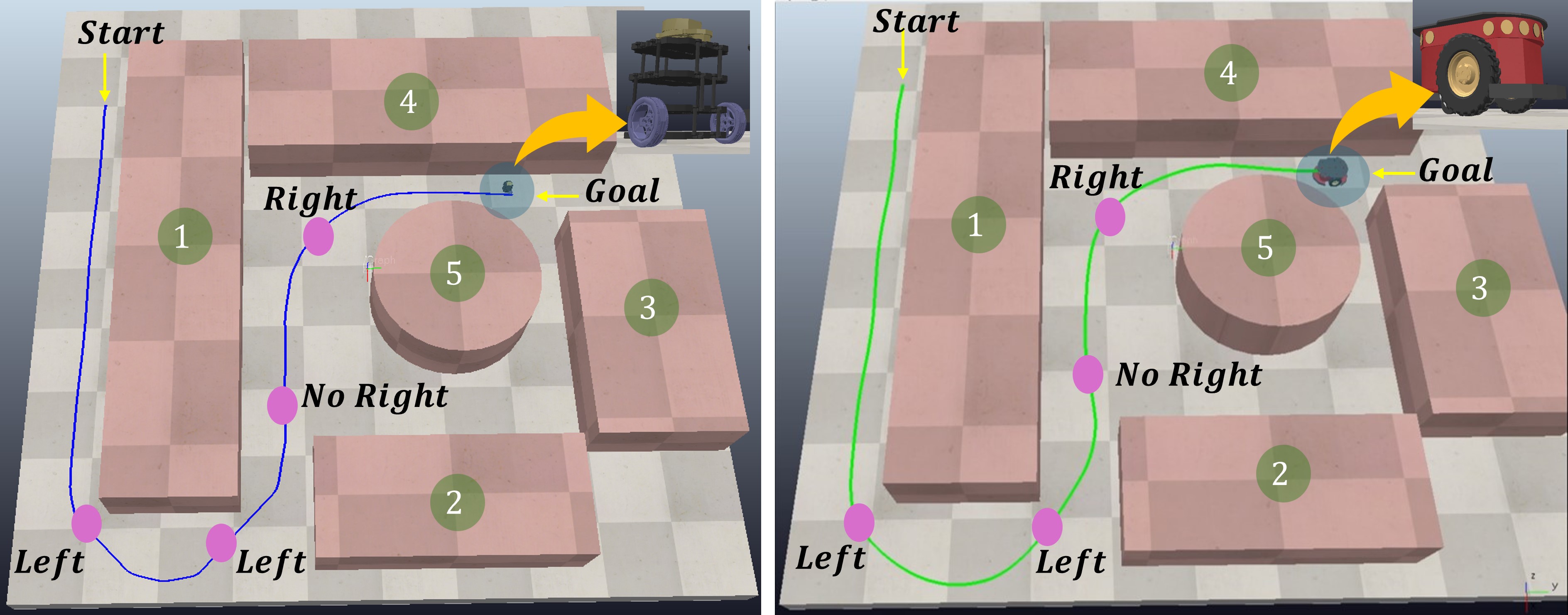}
    \caption{Path generated by LASMP for $\{left, left, no right, right\}$  navigation commands in the random obstacle (RO) scene. The curves in blue (left) and green (right) represent the smoothened path executed by Turtlebot3 and Pioneer 3DX robots.}
    \label{fig:CoppeliaSim_simulation}
\end{figure}
To verify whether the computed paths could be executed on physical robots, we deployed a trajectory-following controller on Pioneer3DX and Turtlebot3 robots. These robots were chosen for their variability in wheel separation distances and wheel radii (Table~\ref{tab:robot_parameters}), two key kinematic parameters influencing control velocities in path-following tasks. Both robots successfully followed the smoothed paths generated by LASMP using spline curve fitting, as shown in Fig.~\ref{fig:CoppeliaSim_simulation}. We also conducted tests using TurtleBot2 in a university building corridor environment, (see Fig.~\ref{fig:real_world}). These experiments validated LASMP’s practicality to handle real-world planning problems.

\begin{table}[!htb]
\begin{center}
\caption{Kinematic parameters of the robotic platforms.}
\begin{tabular}{ |c|c|c| }
\hline
\rowcolor{lightgray} \textbf{Robots} & \textbf{track width} [m] & \textbf{wheel radius} [m]  \\ \hline
Pioneed3DX~\cite{pioneer_robot} & 0.380 & 0.097\\ \hline
Turtlebot3~\cite{turtlebot} & 0.160 & 0.033 \\
  \hline
  TurtleBot2~\cite{turtlebot} & 0.230 & 0.041 \\ \hline
 \end{tabular}
 \label{tab:robot_parameters}
\end{center}
\end{table}
\subsubsection*{\textbf{Discussions about LASMP}}
The convergence of LASMP to a feasible user-instructed path efficiently depends on $\bm{\zeta}(\mathcal{D}_{smp})$ which is determined by the choice of $h$ and $w$. We have considered $h>w$ in all the examples in this work based on the fact that $h$ determines the space along the heading direction of the robot. Larger $h$ will allow the robot to find a state in $\mathcal{D}_{is}$ faster. In the future we would like to determine the optimal values for $h$ and $w$. It can be formally proven that the LASMP holds the probabilistic completeness property like the RRT planner, because of the space limitation, this proof is omitted. The parameter $d$, associated with the ray-casting step, is chosen heuristically based on the average lengths of the aisles of a given environment. In our method, ray-casting is done only in the informed direction of the upcoming turn, reducing computational load. 
\section{Conclusions}\label{sec:conclusions}
LASMP is a hybrid planning method that combines a large language model with sampling-based planning to efficiently generate collision-free paths for mobile robots. By leveraging language-based cues, LASMP focuses sampling on relevant areas, significantly improving efficiency over traditional methods like RRT.
Currently designed for environments with static obstacles, LASMP operates as a global planner. To handle dynamic obstacles, a future extension will incorporate a local planning module. Extensive simulations demonstrated that LASMP consistently outperforms RRT in terms of node generation, query count, path length, and computation time.
LASMP was also validated in real-world tests, proving its practical applicability for smooth and efficient path execution. Future work will focus on dynamic obstacle avoidance and scaling the system for larger environments.
\bibliographystyle{IEEEtran}
\bibliography{thebibliography}

\begin{thebibliography}{10}
\providecommand{\url}[1]{#1}
\csname url@samestyle\endcsname
\providecommand{\newblock}{\relax}
\providecommand{\bibinfo}[2]{#2}
\providecommand{\BIBentrySTDinterwordspacing}{\spaceskip=0pt\relax}
\providecommand{\BIBentryALTinterwordstretchfactor}{4}
\providecommand{\BIBentryALTinterwordspacing}{\spaceskip=\fontdimen2\font plus
\BIBentryALTinterwordstretchfactor\fontdimen3\font minus \fontdimen4\font\relax}
\providecommand{\BIBforeignlanguage}[2]{{%
\expandafter\ifx\csname l@#1\endcsname\relax
\typeout{** WARNING: IEEEtran.bst: No hyphenation pattern has been}%
\typeout{** loaded for the language `#1'. Using the pattern for}%
\typeout{** the default language instead.}%
\else
\language=\csname l@#1\endcsname
\fi
#2}}
\providecommand{\BIBdecl}{\relax}
\BIBdecl

\bibitem{durrant2006simultaneous}
H.~Durrant-Whyte and T.~Bailey, ``Simultaneous localization and mapping: part i,'' \emph{IEEE robotics \& automation magazine}, vol.~13, no.~2, pp. 99--110, 2006.

\bibitem{anderson2018vision}
P.~Anderson, Q.~Wu, D.~Teney, J.~Bruce, M.~Johnson, N.~S{\"u}nderhauf, I.~Reid, S.~Gould, and A.~Van Den~Hengel, ``Vision-and-language navigation: Interpreting visually-grounded navigation instructions in real environments,'' in \emph{Proceedings of the IEEE conference on computer vision and pattern recognition}, 2018, pp. 3674--3683.

\bibitem{deruyttere2019talk2car}
T.~Deruyttere, S.~Vandenhende, D.~Grujicic, L.~Van~Gool, and M.-F. Moens, ``Talk2car: Taking control of your self-driving car,'' \emph{arXiv preprint arXiv:1909.10838}, 2019.

\bibitem{codevilla2018end}
F.~Codevilla, M.~M{\"u}ller, A.~L{\'o}pez, V.~Koltun, and A.~Dosovitskiy, ``End-to-end driving via conditional imitation learning,'' in \emph{2018 IEEE international conference on robotics and automation (ICRA)}.\hskip 1em plus 0.5em minus 0.4em\relax IEEE, 2018, pp. 4693--4700.

\bibitem{tellex2011understanding}
S.~Tellex, T.~Kollar, S.~Dickerson, M.~Walter, A.~Banerjee, S.~Teller, and N.~Roy, ``Understanding natural language commands for robotic navigation and mobile manipulation,'' in \emph{Proceedings of the AAAI Conference on Artificial Intelligence}, vol.~25, no.~1, 2011, pp. 1507--1514.

\bibitem{matuszek2010following}
C.~Matuszek, D.~Fox, and K.~Koscher, ``Following directions using statistical machine translation,'' in \emph{2010 5th ACM/IEEE International Conference on Human-Robot Interaction (HRI)}.\hskip 1em plus 0.5em minus 0.4em\relax IEEE, 2010, pp. 251--258.

\bibitem{shah2023lm}
D.~Shah, B.~Osi{\'n}ski, S.~Levine \emph{et~al.}, ``Lm-nav: Robotic navigation with large pre-trained models of language, vision, and action,'' in \emph{Conference on robot learning}.\hskip 1em plus 0.5em minus 0.4em\relax PMLR, 2023, pp. 492--504.

\bibitem{hu2023planning}
Y.~Hu, J.~Yang, L.~Chen, K.~Li, C.~Sima, X.~Zhu, S.~Chai, S.~Du, T.~Lin, W.~Wang \emph{et~al.}, ``Planning-oriented autonomous driving,'' in \emph{Proceedings of the IEEE/CVF Conference on Computer Vision and Pattern Recognition}, 2023, pp. 17\,853--17\,862.

\bibitem{zhu2017target}
Y.~Zhu, R.~Mottaghi, E.~Kolve, J.~J. Lim, A.~Gupta, L.~Fei-Fei, and A.~Farhadi, ``Target-driven visual navigation in indoor scenes using deep reinforcement learning,'' in \emph{2017 IEEE international conference on robotics and automation (ICRA)}.\hskip 1em plus 0.5em minus 0.4em\relax IEEE, 2017, pp. 3357--3364.

\bibitem{geiger2012we}
A.~Geiger, P.~Lenz, and R.~Urtasun, ``Are we ready for autonomous driving? the kitti vision benchmark suite,'' in \emph{2012 IEEE conference on computer vision and pattern recognition}.\hskip 1em plus 0.5em minus 0.4em\relax IEEE, 2012, pp. 3354--3361.

\bibitem{sriram2019talk}
N.~Sriram, T.~Maniar, J.~Kalyanasundaram, V.~Gandhi, B.~Bhowmick, and K.~M. Krishna, ``Talk to the vehicle: Language conditioned autonomous navigation of self driving cars,'' in \emph{2019 IEEE/RSJ International Conference on Intelligent Robots and Systems (IROS)}.\hskip 1em plus 0.5em minus 0.4em\relax IEEE, 2019, pp. 5284--5290.

\bibitem{rufus2021grounding}
N.~Rufus, K.~Jain, U.~K.~R. Nair, V.~Gandhi, and K.~M. Krishna, ``Grounding linguistic commands to navigable regions,'' in \emph{2021 IEEE/RSJ International Conference on Intelligent Robots and Systems (IROS)}.\hskip 1em plus 0.5em minus 0.4em\relax IEEE, 2021, pp. 8593--8600.

\bibitem{kim2020advisable}
J.~Kim, S.~Moon, A.~Rohrbach, T.~Darrell, and J.~Canny, ``Advisable learning for self-driving vehicles by internalizing observation-to-action rules,'' in \emph{Proceedings of the IEEE/CVF Conference on Computer Vision and Pattern Recognition}, 2020, pp. 9661--9670.

\bibitem{kim2019grounding}
J.~Kim, T.~Misu, Y.-T. Chen, A.~Tawari, and J.~Canny, ``Grounding human-to-vehicle advice for self-driving vehicles,'' in \emph{Proceedings of the IEEE/CVF conference on computer vision and pattern recognition}, 2019, pp. 10\,591--10\,599.

\bibitem{mao2023gpt}
J.~Mao, Y.~Qian, H.~Zhao, and Y.~Wang, ``Gpt-driver: Learning to drive with gpt,'' \emph{arXiv preprint arXiv:2310.01415}, 2023.

\bibitem{sha2023languagempc}
H.~Sha, Y.~Mu, Y.~Jiang, L.~Chen, C.~Xu, P.~Luo, S.~E. Li, M.~Tomizuka, W.~Zhan, and M.~Ding, ``Languagempc: Large language models as decision makers for autonomous driving,'' \emph{arXiv preprint arXiv:2310.03026}, 2023.

\bibitem{ma2024lampilot}
Y.~Ma, C.~Cui, X.~Cao, W.~Ye, P.~Liu, J.~Lu, A.~Abdelraouf, R.~Gupta, K.~Han, A.~Bera \emph{et~al.}, ``Lampilot: An open benchmark dataset for autonomous driving with language model programs,'' in \emph{Proceedings of the IEEE/CVF Conference on Computer Vision and Pattern Recognition}, 2024, pp. 15\,141--15\,151.

\bibitem{hu2019safe}
Z.~Hu, J.~Pan, T.~Fan, R.~Yang, and D.~Manocha, ``Safe navigation with human instructions in complex scenes,'' \emph{IEEE Robotics and Automation Letters}, vol.~4, no.~2, pp. 753--760, 2019.

\bibitem{shah2018follownet}
P.~Shah, M.~Fiser, A.~Faust, J.~C. Kew, and D.~Hakkani-Tur, ``Follownet: Robot navigation by following natural language directions with deep reinforcement learning,'' \emph{arXiv:1805.06150}, 2018.

\bibitem{shalev2016sample}
S.~Shalev-Shwartz and A.~Shashua, ``On the sample complexity of end-to-end training vs. semantic abstraction training,'' \emph{arXiv preprint arXiv:1604.06915}, 2016.

\bibitem{rasouli2017effect}
A.~Rasouli and J.~K. Tsotsos, ``The effect of color space selection on detectability and discriminability of colored objects,'' \emph{arXiv preprint arXiv:1702.05421}, 2017.

\bibitem{jain2023ground}
K.~Jain, V.~Chhangani, A.~Tiwari, K.~M. Krishna, and V.~Gandhi, ``Ground then navigate: Language-guided navigation in dynamic scenes,'' in \emph{2023 IEEE International Conference on Robotics and Automation (ICRA)}.\hskip 1em plus 0.5em minus 0.4em\relax IEEE, 2023, pp. 4113--4120.

\bibitem{kuo2018deep}
Y.-L. Kuo, A.~Barbu, and B.~Katz, ``Deep sequential models for sampling-based planning,'' in \emph{2018 IEEE/RSJ International Conference on Intelligent Robots and Systems (IROS)}.\hskip 1em plus 0.5em minus 0.4em\relax IEEE, 2018, pp. 6490--6497.

\bibitem{kuo2020deep}
Y.-L. Kuo, B.~Katz, and A.~Barbu, ``Deep compositional robotic planners that follow natural language commands,'' in \emph{2020 IEEE international conference on robotics and automation (ICRA)}.\hskip 1em plus 0.5em minus 0.4em\relax IEEE, 2020, pp. 4906--4912.

\bibitem{lavalle2001randomized}
S.~M. LaValle and J.~J. Kuffner~Jr, ``Randomized kinodynamic planning,'' \emph{The international journal of robotics research}, vol.~20, no.~5, pp. 378--400, 2001.

\bibitem{kuffner2000rrt}
J.~J. Kuffner and S.~M. LaValle, ``Rrt-connect: An efficient approach to single-query path planning,'' in \emph{Proceedings 2000 ICRA. Millennium Conference. IEEE International Conference on Robotics and Automation. Symposia Proceedings (Cat. No. 00CH37065)}, vol.~2.\hskip 1em plus 0.5em minus 0.4em\relax IEEE, 2000, pp. 995--1001.

\bibitem{karaman2011sampling}
S.~Karaman and E.~Frazzoli, ``Sampling-based algorithms for optimal motion planning,'' \emph{The international journal of robotics research}, vol.~30, no.~7, pp. 846--894, 2011.

\bibitem{zucker2007multipartite}
M.~Zucker, J.~Kuffner, and M.~Branicky, ``Multipartite rrts for rapid replanning in dynamic environments,'' in \emph{Proceedings 2007 IEEE International Conference on Robotics and Automation}.\hskip 1em plus 0.5em minus 0.4em\relax IEEE, 2007, pp. 1603--1609.

\bibitem{gammell2014informed}
J.~D. Gammell, S.~S. Srinivasa, and T.~D. Barfoot, ``Informed rrt: Optimal sampling-based path planning focused via direct sampling of an admissible ellipsoidal heuristic,'' in \emph{2014 IEEE/RSJ international conference on intelligent robots and systems}.\hskip 1em plus 0.5em minus 0.4em\relax IEEE, 2014, pp. 2997--3004.

\bibitem{kostavelis2015semantic}
I.~Kostavelis and A.~Gasteratos, ``Semantic mapping for mobile robotics tasks: A survey,'' \emph{Robotics and Autonomous Systems}, vol.~66, pp. 86--103, 2015.

\bibitem{mei2016listen}
H.~Mei, M.~Bansal, and M.~Walter, ``Listen, attend, and walk: Neural mapping of navigational instructions to action sequences,'' in \emph{Proceedings of the AAAI Conference on Artificial Intelligence}, vol.~30, no.~1, 2016.

\bibitem{gong2023listen}
Y.~Gong, H.~Luo, A.~H. Liu, L.~Karlinsky, and J.~Glass, ``Listen, think, and understand,'' \emph{arXiv preprint arXiv:2305.10790}, 2023.

\bibitem{zhang2023speechgpt}
D.~Zhang, S.~Li, X.~Zhang, J.~Zhan, P.~Wang, Y.~Zhou, and X.~Qiu, ``Speechgpt: Empowering large language models with intrinsic cross-modal conversational abilities,'' \emph{arXiv preprint arXiv:2305.11000}, 2023.

\bibitem{xu2024drivegpt4}
Z.~Xu, Y.~Zhang, E.~Xie, Z.~Zhao, Y.~Guo, K.-Y.~K. Wong, Z.~Li, and H.~Zhao, ``Drivegpt4: Interpretable end-to-end autonomous driving via large language model,'' \emph{IEEE Robotics and Automation Letters}, 2024.

\bibitem{chen2023videollm}
G.~Chen, Y.-D. Zheng, J.~Wang, J.~Xu, Y.~Huang, J.~Pan, Y.~Wang, Y.~Wang, Y.~Qiao, T.~Lu \emph{et~al.}, ``Videollm: Modeling video sequence with large language models,'' \emph{arXiv preprint arXiv:2305.13292}, 2023.

\bibitem{guo2023point}
Z.~Guo, R.~Zhang, X.~Zhu, Y.~Tang, X.~Ma, J.~Han, K.~Chen, P.~Gao, X.~Li, H.~Li \emph{et~al.}, ``Point-bind \& point-llm: Aligning point cloud with multi-modality for 3d understanding, generation, and instruction following,'' \emph{arXiv preprint arXiv:2309.00615}, 2023.

\bibitem{xu2023pointllm}
R.~Xu, X.~Wang, T.~Wang, Y.~Chen, J.~Pang, and D.~Lin, ``Pointllm: Empowering large language models to understand point clouds,'' \emph{arXiv preprint arXiv:2308.16911}, 2023.

\bibitem{kollar2013generalized}
T.~Kollar, S.~Tellex, M.~R. Walter, A.~Huang, A.~Bachrach, S.~Hemachandra, E.~Brunskill, A.~Banerjee, D.~Roy, S.~Teller \emph{et~al.}, ``Generalized grounding graphs: A probabilistic framework for understanding grounded language,'' \emph{Journal of Artificial Intelligence Research}, pp. 1--35, 2013.

\bibitem{park2017generating}
J.~S. Park, B.~Jia, M.~Bansal, and D.~Manocha, ``Generating realtime motion plans from complex natural language commands using dynamic grounding graphs,'' \emph{arXiv preprint arXiv:1707.02387}, 2017.

\bibitem{howard2014natural}
T.~M. Howard, S.~Tellex, and N.~Roy, ``A natural language planner interface for mobile manipulators,'' in \emph{2014 IEEE International Conference on Robotics and Automation (ICRA)}.\hskip 1em plus 0.5em minus 0.4em\relax IEEE, 2014, pp. 6652--6659.

\bibitem{chen2024driving}
L.~Chen, O.~Sinavski, J.~H{\"u}nermann, A.~Karnsund, A.~J. Willmott, D.~Birch, D.~Maund, and J.~Shotton, ``Driving with llms: Fusing object-level vector modality for explainable autonomous driving,'' in \emph{2024 IEEE International Conference on Robotics and Automation (ICRA)}.\hskip 1em plus 0.5em minus 0.4em\relax IEEE, 2024, pp. 14\,093--14\,100.

\bibitem{xie2023language}
A.~Xie, Y.~Lee, P.~Abbeel, and S.~James, ``Language-conditioned path planning,'' in \emph{Conference on Robot Learning}.\hskip 1em plus 0.5em minus 0.4em\relax PMLR, 2023, pp. 3384--3396.

\bibitem{radford2023robust}
A.~Radford, J.~W. Kim, T.~Xu, G.~Brockman, C.~McLeavey, and I.~Sutskever, ``Robust speech recognition via large-scale weak supervision,'' in \emph{International conference on machine learning}.\hskip 1em plus 0.5em minus 0.4em\relax PMLR, 2023, pp. 28\,492--28\,518.

\bibitem{Honnibal_spaCy_Industrial-strength_Natural_2020}
M.~Honnibal, I.~Montani, S.~Van~Landeghem, and A.~Boyd, ``{spaCy: Industrial-strength Natural Language Processing in Python},'' 2020.

\bibitem{devlin2018bert}
J.~Devlin, M.-W. Chang, K.~Lee, and K.~Toutanova, ``Bert: Pre-training of deep bidirectional transformers for language understanding,'' \emph{arXiv preprint arXiv:1810.04805}, 2018.

\bibitem{liu2019roberta}
Y.~Liu, M.~Ott, N.~Goyal, J.~Du, M.~Joshi, D.~Chen, O.~Levy, M.~Lewis, L.~Zettlemoyer, and V.~Stoyanov, ``Roberta: A robustly optimized bert pretraining approach,'' \emph{arXiv:1907.11692}, 2019.

\bibitem{lavalle2006planning}
S.~M. LaValle, \emph{Planning algorithms}.\hskip 1em plus 0.5em minus 0.4em\relax Cambridge university press, 2006.

\bibitem{lozano1990spatial}
T.~Lozano-Perez, \emph{Spatial planning: A configuration space approach}.\hskip 1em plus 0.5em minus 0.4em\relax Springer, 1990.

\bibitem{anshelevich2000deformable}
E.~Anshelevich, S.~Owens, F.~Lamiraux, and L.~E. Kavraki, ``Deformable volumes in path planning applications,'' in \emph{Proceedings 2000 ICRA. Millennium Conference. IEEE International Conference on Robotics and Automation. Symposia Proceedings (Cat. No. 00CH37065)}, vol.~3.\hskip 1em plus 0.5em minus 0.4em\relax IEEE, 2000, pp. 2290--2295.

\bibitem{sinha2023oc3}
A.~Sinha, R.~Laha, and N.~Chakraborty, ``Oc3: A reactive velocity level motion planner with complementarity constraint-based obstacle avoidance for mobile robots,'' in \emph{2023 IEEE 19th International Conference on Automation Science and Engineering (CASE)}.\hskip 1em plus 0.5em minus 0.4em\relax IEEE, 2023, pp. 1--8.

\bibitem{6696520}
E.~{Rohmer}, S.~P.~N. {Singh}, and M.~{Freese}, ``V-rep: A versatile and scalable robot simulation framework,'' in \emph{2013 IEEE/RSJ International Conference on Intelligent Robots and Systems}, 2013, pp. 1321--1326.

\bibitem{turtlebot}
``Turtlebot,'' \url{https://www.turtlebot.com/}, 2024.

\bibitem{pioneer_robot}
``Pioneer robot,'' \url{https://robots.ros.org/pioneer-3-dx/}, 2024.

\bibitem{the_dataset}
``Lasmp dataset,'' \url{https://github.com/LASMP23/LASMP}, 2024.

\bibitem{openai2024chatgpt}
\BIBentryALTinterwordspacing
OpenAI, ``Chatgpt (september 11 version),'' 2024, accessed: 2024-09-11. [Online]. Available: \url{https://chat.openai.com/}
\BIBentrySTDinterwordspacing

\bibitem{spacy_annotation_tool}
``Spacy annotation tool,'' \url{https://tecoholic.github.io/ner-annotator/}.

\bibitem{vaswani2017attention}
A.~Vaswani, ``Attention is all you need,'' \emph{Advances in Neural Information Processing Systems}, 2017.

\end{thebibliography}
\end{document}